\documentclass[journal]{IEEEtran}
\ifCLASSINFOpdf
\else
\fi
\usepackage{graphicx}
\usepackage{subfigure}
\usepackage{multirow}
\usepackage{color}
\usepackage{ulem}

\begin{document}
%
\title{Dual-Sampling Attention Network for Diagnosis of COVID-19 from Community Acquired Pneumonia}
%
%
%

\author{Xi Ouyang$^{\dag}$, Jiayu Huo$^{\dag}$, Liming Xia$^{\dag}$, Fei Shan$^{\dag}$, Jun Liu$^{\dag}$, Zhanhao Mo$^{\dag}$, Fuhua Yan$^{\dag}$, Zhongxiang Ding$^{\dag}$, Qi Yang$^{\dag}$, Bin Song$^{\dag}$, Feng Shi, Huan Yuan, Ying Wei, Xiaohuan Cao, Yaozong Gao, Dijia Wu, Qian Wang$^{*}$, Dinggang Shen$^{*}$
\thanks{${\dag}$ X. Ouyang, J. Huo, L. Xia, F. Shan, J. Liu, Z. Mo, F. Yan, Z. Ding, Q. Yang, and B. Song contributed equally to this work.}
\thanks{$*$ Corresponding authors: Q. Wang (wang.qian@sjtu.edu.cn) and D. Shen (Dinggang.Shen@gmail.com).}
\thanks{X. Ouyang, J. Huo and Q. Wang are with the Institute for Medical Imaging Technology, School of Biomedical Engineering, Shanghai Jiao Tong University, Shanghai, China. X. Ouyang and J. Huo are interns at Shanghai United Imaging Intelligence Co. during this work. (e-mail: \{xi.ouyang,  jiayu.huo, wang.qian\}@sjtu.edu.cn). }
\thanks{L. Xia is with the Department of Radiology, Tongji Hospital, Tongji Medical College, Huazhong University of Science and Technology, Wuhan, Hubei, China. (e-mail: xialiming2017@outlook.com).}
\thanks{F. Shan is with the Department of Radiology, Shanghai Public Health Clinical Center, Fudan University, Shanghai, China. (e-mail: shanfei\_2901@163.com).}
\thanks{J. Liu is with the Department of Radiology, The Second Xiangya Hospital, Central South University, Changsha, Hunan Province, China, and is also with the Department of Radiology Quality Control Center, Changsha, Hunan Province, China. (e-mail: junliu123@csu.edu.cn).}
\thanks{Z. Mo is with the Department of Radiology, China-Japan Union Hospital of Jilin University, Changchun, China. (e-mail: mozhanhao@jlu.edu.cn).}
\thanks{F. Yan is with the Department of Radiology, Ruijin Hospital, Shanghai Jiao Tong University School of Medicine, Shanghai, China. (e-mail: yfh11655@rjh.com.cn).}
\thanks{Z. Ding is with the Department of Radiology, Affiliated Hangzhou First People’s Hospital, Zhejiang University School of Medicine, Hangzhou, Zhejiang, China. (e-mail: hangzhoudzx73@126.com).}
\thanks{Q. Yang is with the Beijing Chaoyang hospital, Capital Medical University. (e-mail: yangyangqiqi@gmail.com).}
\thanks{B. Song is with the Department of Radiology, Sichuan University West China Hospital, Chengdu, China. (e-mail: anicesong@vip.sina.com).}
\thanks{F. Shi, H. Yuan, Y. Wei, X. Cao, Y. Gao, D. Wu and D. Shen are with the Department of Research and Development, Shanghai United Imaging Intelligence Co., Ltd., Shanghai, China. (e-mail: \{feng.shi, huan.yuan, ying.wei, xiaohuan.cao, yaozong.gao, dijia.wu\}@united-imaging.com, Dinggang.Shen@gmail.com).}
}

%
%

\markboth{}%
{Shell \MakeLowercase{\textit{et al.}}: Bare Demo of IEEEtran.cls for IEEE Journals}
%



\maketitle

\begin{abstract}
The coronavirus disease (COVID-19) is rapidly spreading all over the world, and has infected more than 1,436,000 people in more than 200 countries and territories as of April 9, 2020. Detecting COVID-19 at early stage is essential to deliver proper healthcare to the patients and also to protect the uninfected population. To this end, we develop a dual-sampling attention network to automatically diagnose COVID-19 from the community acquired pneumonia (CAP) in chest computed tomography (CT). In particular, we propose a novel online attention module with a 3D convolutional network (CNN) to focus on the infection regions in lungs when making decisions of diagnoses. Note that there exists imbalanced distribution of the sizes of the infection regions between COVID-19 and CAP, partially due to fast progress of COVID-19 after symptom onset. Therefore, we develop a dual-sampling strategy to mitigate the imbalanced learning. Our method is evaluated (to our best knowledge) upon the largest multi-center CT data for COVID-19 from 8 hospitals. In the training-validation stage, we collect 2186 CT scans from 1588 patients for a 5-fold cross-validation. In the testing stage, we employ another independent large-scale testing dataset including 2796 CT scans from 2057 patients. Results show that our algorithm can identify the COVID-19 images with the area under the receiver operating characteristic curve (AUC) value of 0.944, accuracy of 87.5\%, sensitivity of 86.9\%, specificity of 90.1\%, and F1-score of 82.0\%. With this performance, the proposed algorithm could potentially aid radiologists with COVID-19 diagnosis from CAP, especially in the early stage of the COVID-19 outbreak.
\end{abstract}


%
\IEEEpeerreviewmaketitle

\section{Introduction}
%
%
%
%
\IEEEPARstart{T}{he} disease caused by the novel coronavirus, or Coronavirus Disease 2019 (COVID-19) is quickly spreading globally. It has infected more than 1,436,000 people in more than 200 countries and territories as of April 9, 2020 \cite{world2020coronavirus80}. On February 12, 2020, the World Health Organization (WHO) officially named the disease caused by the novel coronavirus as Coronavirus Disease 2019 (COVID-19) \cite{world2020director}. Now, the number of COVID-19 patients, is dramatically increasing every day around the world \cite{world2020coronavirus}. Compared with the prior Severe Acute Respiratory Syndrome (SARS) and Middle East Respiratory Syndrome (MERS), COVID-19 has spread to more places and caused more deaths, despite its relatively lower fatality rate \cite{wu2020characteristics,mahase2020coronavirus}. Considering the pandemic of COVID-19, it is important to detect COVID-19 early, which could facilitate the slowdown of viral transmission and thus disease containment.

In clinics, real-time reverse-transcription–polymerase-chain-reaction (RT-PCR) is the golden standard to make a definitive diagnosis of COVID-19 infection \cite{zu2020coronavirus}. However, the high false negative rate \cite{chan2020familial} and unavailability of RT-PCR assay in the early stage of an outbreak may delay the identification of potential patients. Due to the highly contagious nature of the virus, it then constitutes a high risk for infecting a larger population. At the same time, thoracic computed tomography (CT) is relatively easy to perform and can produce fast diagnosis \cite{ai2020correlation}. For example, almost all COVID-19 patients have some typical radiographic features in chest CT, including ground-glass opacities (GGO), multifocal patchy consolidation, and/or interstitial changes with a peripheral distribution \cite{chung2020ct}. Thus chest CT has been recommended as a major tool for clinical diagnosis especially in the hard-hit region such as Hubei, China \cite{zu2020coronavirus}. Considering the need of high-throughput screening by chest CT and the workload for radiologists especially in the outbreak, we design a deep-learning-based method to automatically diagnose COVID-19 infection from the community acquired pneumonia (CAP) infection.

\begin{figure}[!t]
\centering
\includegraphics[width=8.6cm, height=7.5cm]{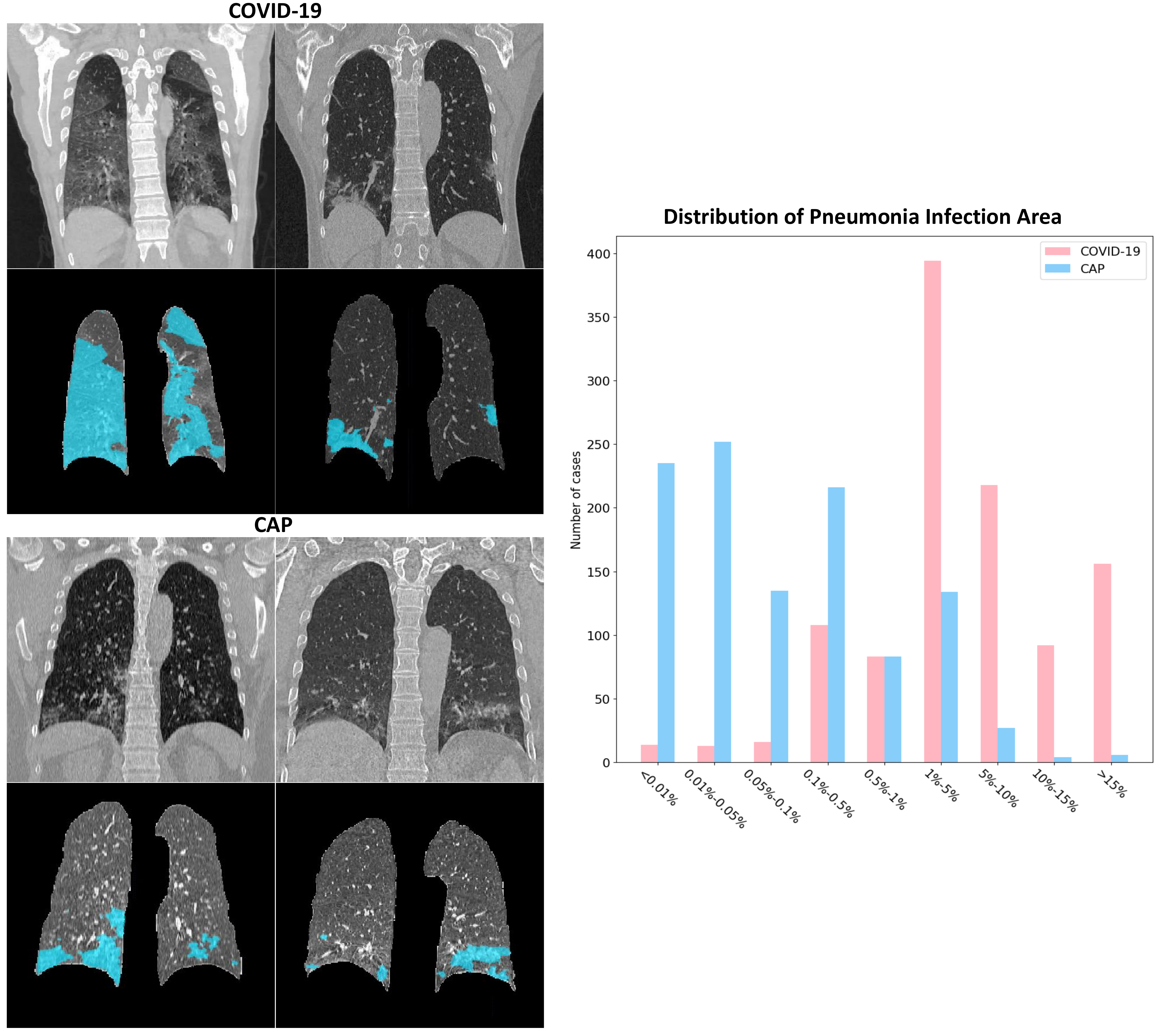}
\caption{Examples of CT images and infection segmentations of two COVID-19 patients (upper left) and two CAP patients (bottom left), and the size distribution of the infection regions of COVID-19 and CAP in our training-validation set (right). The segmentation results of the lungs and infection regions are obtained from an established VB-Net toolkit \cite{shan+2020lung}. The sizes of the infection regions are denoted by the volume ratio of the segmented infection regions and the whole lung. Compared with CAP, the COVID-19 cases tend to have more severe infections in terms of the infection region sizes.} 
\label{fig:intro}
\end{figure}

With the development of deep learning \cite{lecun2015deep,krizhevsky2012imagenet,he2016deep,huang2017densely,nie2016estimating}, the technology has a wide range of applications in medical image processing, including disease diagnosis \cite{wang2017chestx}, and organ segmentation \cite{ronneberger2015u}, etc. Convolutional neural network (CNN) \cite{lecun1989backpropagation}, one of the most representative deep learning technology, has been applied to reading and analyzing CT images in many recent studies \cite{pang2019automatic,park2019lung}. For example, Koichiro et. al. use CNN for differentiation of liver masses on dynamic contrast agent–enhanced CT images \cite{yasaka2018deep}. Also, some studies focus on the diagnoses of lung diseases in chest CT, e.g., pulmonary nodules \cite{huang2018added,ardila2019end} and pulmonary tuberculosis \cite{lakhani2017deep}. Although deep learning has achieved remarkable performance for abnormality diagnoses of medical images \cite{wang2017chestx,irvin2019chexpert,cruz2013deep}, physicians have concerns especially in the lack of model interpretability and understanding \cite{zhang2018visual}, which is important for the diagnosis of COVID-19. To provide more insight for model decisions, the class activation mapping (CAM) \cite{zhou2016learning} and gradient-weighted class activation mapping (Grad-CAM) \cite{selvaraju2017grad} methods have been proposed to produce localization heatmaps highlighting important regions that are closely associated with predicted results. 

In this study, we propose a dual-sampling attention network to classify the COVID-19 and CAP infection. To focus on the lung, our method leverages a lung mask to suppress image context of none-lung regions in chest CT. At the same time, we refine the attention of the deep learning model through an online mechanism, in order to better focus on the infection regions in the lung. In this way, the model facilitates interpreting and explaining the evidence for the automatic diagnosis of COVID-19. The experimental results also demonstrate that the proposed online attention refinement can effectively improve classification performance. 

In our work, an important observation is that COVID-19 cases usually have more severe infection than CAP cases \cite{shi2020large}, although some COVID-19 cases and CAP cases do have similar infection sizes. To illustrate it, we use an established VB-Net toolkit \cite{shan+2020lung} to automatically segment lungs and pneumonia infection regions on all the cases in our training-validation (TV) set (with details of our TV set provided in Section \ref{experiment}), and show the distribution of the ratios between the infection regions and lungs in Fig. \ref{fig:intro}. We can see the imbalanced distribution of the infection size ratios in both COVID-19 and CAP data. In this situation, the conventional uniform sampling on the entire dataset to train the network could lead to unsatisfactory diagnosis performance, especially concerning the limited cases of COVID-19 with small infections and also the limited cases of CAP with large infections. To this end, we train the second network with the size-balanced sampling strategy, by sampling more cases of COVID-19 with small infections and also more cases of CAP with large infections within mini-batches. Finally, we apply ensemble learning to integrate the networks of uniform sampling and size-balanced sampling to get the final diagnosis results, by following the dual-sampling strategy.

As a summary, the contributions of our work are in three-fold:
\begin{itemize}
\item We propose an online module to utilize the segmented pneumonia infection regions to refine the attention for the network. This ensures the network to focus on the infection regions and increase the adoption of visual attention for model interpretability and explainability.
\item We propose a dual-sampling strategy to train the network, which further alleviates the imbalanced distribution of the sizes of pneumonia infection regions.
\item To our knowledge, we have used the largest multi-center CT data in the world for evaluating automatic COVID-19 diagnosis. In particular, we conduct extensive cross-validations in a TV dataset of 2186 CT scans from  1588 patients. Moreover, to better evaluate the performance and generalization ability of the proposed method, a large independent testing set of 2796 CT scans from 2057 patients is also used. Experimental results demonstrate that our algorithm is able to identify the COVID-19 images with the area under the receiver operating characteristic curve (AUC) value of 0.944, accuracy of 87.5\%, sensitivity of 86.9\%, specificity of 90.1\%, and F1-score of 82.0\%.
\end{itemize}

\section{RELATED WORKS}
\subsection{Computer-Assisted Pneumonia Diagnosis}
Chest X-ray (CXR) is one of the firstline imaging modality to diagnose pneumonia, which manifests as increased opacity \cite{franquet2018imaging}. The CNN networks have been successfully applied to pneumonia diagnosis in CXR images \cite{wang2017chestx,rajpurkar2017chexnet}. As the release of the Radiological Society of North America (RSNA) pneumonia detection challenge \cite{challenge2018radiological} dataset, object detection methods (i.e., RetinaNet \cite{lin2017focal} and Mask R-CNN \cite{he2017mask}) have been used for pneumonia localization in CXR images. At the same time, CT has been used as a standard procedure in the diagnosis of lung diseases \cite{wielputz2014radiological}. An automated classification method has been proposed to use regional volumetric texture analysis for usual interstitial pneumonia diagnosis in high-resolution CT \cite{depeursinge2015automated}. For COVID-19, GGO and consolidation along the subpleural area of the lung are the typical radiographic features of COVID-19 patients \cite{chung2020ct}. Chest CT, especially high-resolution CT, can detect small areas of ground glass opacity (GGO) \cite{macmahon2017guidelines}.


Some recent works have focused on the COVID-19 diagnosis from other pneumonia in CT images \cite{wang2020deep,xu2020deep,song2020deep}. It requires the chest CT images to identify some typical features, including GGO, multifocal patchy consolidation, and/or interstitial changes with a peripheral distribution \cite{chung2020ct}. Wang et al. \cite{wang2020deep} propose a 2D CNN network to classify between COVID-19 and other viral pneumonia based on manually delineated regions. Xu et al. \cite{xu2020deep} use a V-Net model to segment the infection region and apply a ResNet18 network for the classification. Ying et al. \cite{song2020deep} use a ResNet50 network to process all the slices of each 3D chest CT images to form the final prediction for each CT images. However, all these methods are evaluated in small datasets. In this paper, we have collected 4982 CT scans from 3645 patients, provided by 8 collaborative hospitals. To our best knowledge, it is the largest multi-center dataset for COVID-19 till now, which can prove the effectiveness of the method.  

Note that, in the context of pneumonia diagnosis, lung segmentation is often an essential preprocessing step in analyzing chest CT images to assess pneumonia. In the literature, Alom et al. \cite{alom2018recurrent} utilize U-net, residual network and recurrent CNN for lung lesion segmentation. A convolutional-deconvolutional capsule network has also been proposed for pathological lung segmentation in CT images. In this paper, we use an established VB-Net toolkit for lung segmentation, which has been reported with high Dice similarity coefficient of $>$ 98\% in evaluation \cite{shan+2020lung}. Also, this VB-Net toolkit achieves Dice similarity coefficient of 92\% between automatically and manually delineated pneumonia infection regions, showing the state-of-the-art performance \cite{9069255}. For more related works, a recent review paper of automatic segmentation methods on COVID-19 could be found in \cite{9069255}.

\begin{figure*}[!t]
\centering
\includegraphics[width=0.95\textwidth]{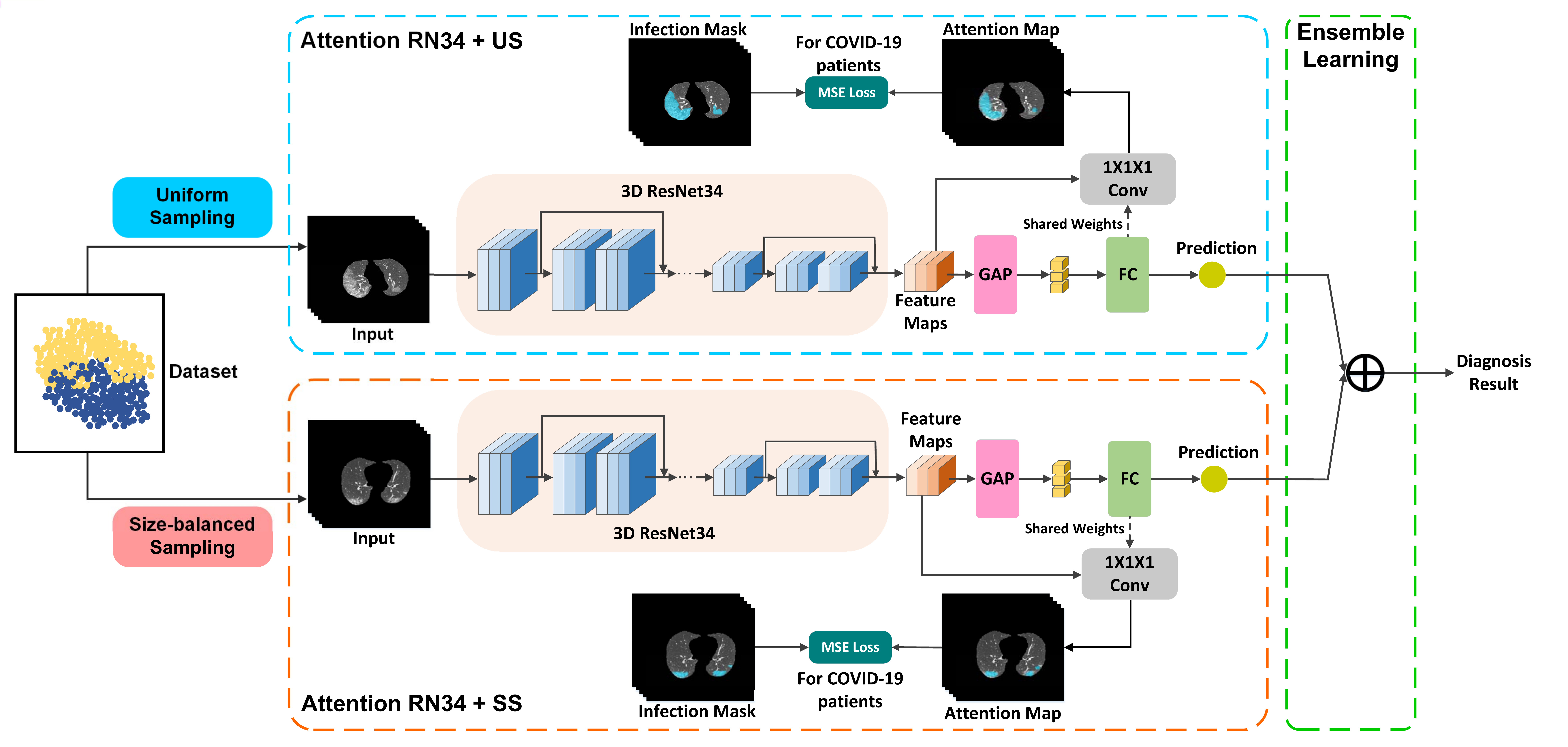}
\caption{Illustration of the pipeline of the proposed method, including two steps. 1) We train two 3D ResNet34 networks \cite{hara2018can} with different sampling strategies. Also, the online attention mechanism generates attention maps during training, which refer to the segmented infection regions to refine the attention localization. 2) We use the ensemble learning to integrate predictions from the two trained networks. In this figure, ``Attention RN34 + US" means the 3D ResNet34 (RN34) with attention module and uniform sampling (US) strategy, while ``Attention RN34 + SS" means the 3D ResNet34 with attention module and size-balanced sampling (SS) strategy. ``GAP" indicates the global average pooling layer, and ``FC" indicates the fully connected layer. ``$1\times1\times1$ Conv" refers to the convolutional layer with $1\times1\times1$ kernel, and takes the parameters from the fully connected layer as the kernel weights. ``MSE Loss" refers to the mean square error function.} 
\label{fig:framework}
\end{figure*}

\subsection{Class Re-sampling Strategies}
For network training in the datasets with long-tailed data distribution, there exist some problems for the universal paradigm to sample the entire dataset uniformly \cite{van2017devil}. In such datasets, some classes contain relatively few samples. The information of these cases may be ignored by the network if applying uniform sampling. To address this, some class re-sampling strategies have been proposed in the literature \cite{zhou2019bbn,buda2018systematic,shen2016relay,he2009learning,japkowicz2002class}. The aim of these methods is to adjust the numbers of the examples from different classes within mini-batches, which achieves better performance on long-tailed dataset. Generally, class re-sampling strategies could be categorized into two groups, i.e., over-sampling by repeating data for minority classes \cite{zhou2019bbn,buda2018systematic,shen2016relay} and under-sampling by randomly removing samples to make the number of each class to be equal \cite{buda2018systematic,he2009learning,japkowicz2002class}. The COVID-19 data is hard to collect and precious, so abandoning data is not a good choice. In this study, we adapt the over-sampling strategies \cite{zhou2019bbn} on the COVID-19 with small infections and also CAP with large infections to form a size-balanced sampling method, which can better balance the distribution of the infection regions of COVID-19 and CAP cases within mini-batches. However, over-sampling may lead to over-fitting upon these minority classes \cite{cui2019class,chawla2002smote}. We thus propose the dual-sampling strategy to integrate results from the two networks trained with uniform sampling and size-balanced sampling, respectively.

\begin{figure}
\centering
\includegraphics[width=7.4cm, height=3.8cm]{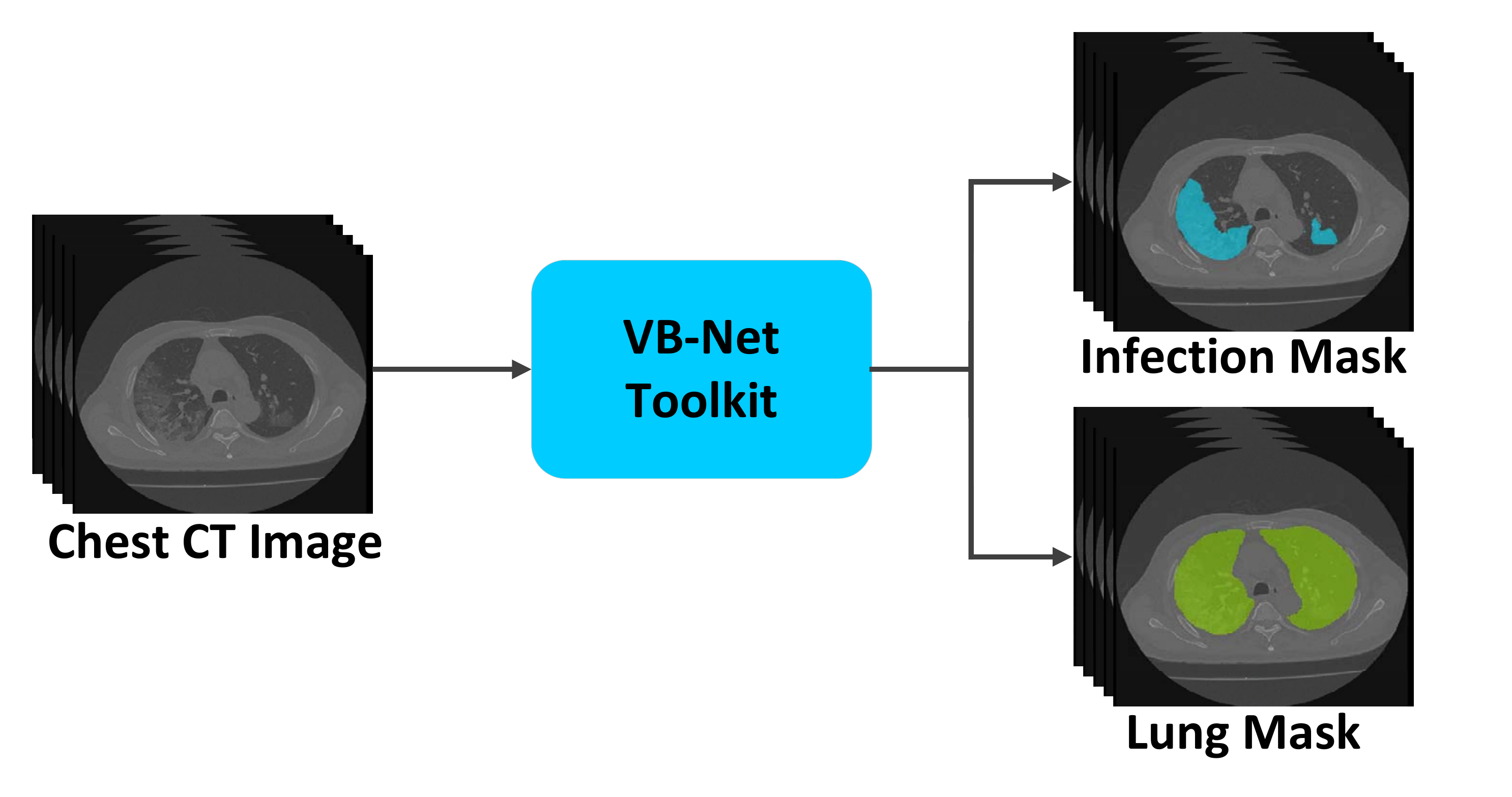}
\caption{The pneumonia infection region (upper right) and the lung segmentation (bottom right) from the VB-Net toolkit \cite{shan+2020lung}.} 
\label{fig:vbnet}
\end{figure}

\subsection{Attention Mechanism}
Attention mechanism has been widely used in many deep networks, and can be roughly divided into two types: 1) activation-based attention \cite{wang2018non,fu2019dual,hu2018squeeze} and 2) gradient-based attention \cite{zhou2016learning,selvaraju2017grad}. The activation-based attention usually serves as an inserted module to refine the hidden feature maps during the training, which can make the network to focus on the important regions. For the activation-based attention, the channel-wise attention assigns weights to each channel in the feature maps \cite{hu2018squeeze} while the position-wise attention produces heatmaps of importance for each pixel of the feature maps \cite{wang2018non,fu2019dual}. The most common gradient-based attention methods are CAM \cite{zhou2016learning} and Grad-CAM \cite{selvaraju2017grad}, which reveal the important regions influencing the network prediction. These methods are normally conducted offline and provide a pattern of model interpretability during the inference stage. Recently, some studies \cite{fukui2019attention,li2018tell} argue that the gradient-based methods can be developed as an online module during the training for better localization. In this study, we extend the gradient-based attention to composing an online trainable component and the scenario of 3D input. The proposed attention module utilizes the segmented pneumonia infection regions to ensure that the network can make decisions based on these infection regions.

\section{METHOD}
The overall framework is shown in Fig. \ref{fig:framework}. The input for the network is the 3D CT images masked in lungs only. We use an established VB-Net toolkit \cite{shan+2020lung} to segment the lungs for all CT images, and perform auto-contouring of possible infection regions as shown in Fig. \ref{fig:vbnet}. 
The VB-Net toolkit is a modified network that combines V-Net \cite{milletari2016v} with bottleneck layers to reduce and integrate feature map channels. The toolkit is capable of segmenting the infected regions as well as the lung fields, achieving Dice similarity coefficient of 92\% between automatically and manually delineated infection regions \cite{shan+2020lung}.
By labeling all voxels within the segmented regions to $1$, and the rest part to $0$, we can get the corresponding lung mask and then input image by masking the original CT image with the corresponding lung mask.

As shown in Fig. \ref{fig:framework}, the training pipeline of our method consists of two stages: 1) using different sampling strategies to train two 3D ResNet34 models \cite{hara2018can} with the online attention module; 2) training an ensemble learning layer to integrate the predictions from the two models. The details of our method are introduced in the following sections.

\subsection{Network}
We use the 3D ResNet34 architecture \cite{hara2018can} as the backbone network. It is the 3D extended version of residual network \cite{he2016deep}, which uses the 3D kernels in all the convolutional layers. In 3D ResNet34, we set the stride of each dimension as $1$ in the last residual block instead of $2$. This makes the resolution of the feature maps before the global average pooling (GAP) \cite{lin2013network} operation into $1/16$ of the input CT image in each dimension. Compared with the case of downsampling the input image by a factor of 32 in each dimension in the original 3D ResNet34, it can greatly improve the quality of the generated attention maps based on higher-resolution feature maps.

\subsection{Online attention module}
To exhaustively learn all features that are important for classification, and also to produce the corresponding attention maps, we use an online attention mechanism of 3D class activation mapping (CAM). The key idea of CAM \cite{zhou2016learning,selvaraju2017grad,fukui2019attention} is to back-propagate weights of the fully-connected layer onto the convolutional feature maps for generating the attention maps. In this study, we extend this offline operation to become an online trainable component for the scenario of 3D input. Let $f$ denote the feature maps before the GAP operation and also $w$ denote the weight matrix of the fully-connected layer. To make our attention generation procedure trainable, we use $w$ as the kernel of a $1\times1\times1$ convolution layer and apply a ReLU layer \cite{nair2010rectified} to generate the attention feature map $A$ as:
\begin{normalsize}
\begin{equation}
\centering
\label{eq:cam}
A = {\rm{ReLU}}\left( {{\rm{conv}}\left( {f,w} \right)} \right),
\end{equation}
\end{normalsize}
where $A$ has the shape $X \times Y \times Z$, and $X,Y,Z$ is $1/16$ of corresponding size of the input CT images. Given the attention feature map $A$, we first upsample it to the input image size, then normalize it to have intensity values between $0$ and $1$, and finally perform sigmoid for soft masking \cite{li2018tell}, as follows:
\begin{normalsize}
\begin{equation}
\label{eq:thresholding}
T(A) = \frac{1}{{1 + \exp ( - \alpha (A - \beta ))}},
\end{equation}
\end{normalsize}
where values of $\alpha$ and $\beta$ are set to $100$ and $0.4$ respectively. $T(A)$ is the generated attention map of this online attention module, where $A$ is defined in Eq. \ref{eq:cam}. During the training, the parameters in the $1\times1\times1$ convolution layer are always copied from the fully-connected layer and only updated by the binary cross entropy (BCE) loss for the classification task.

\subsection{Size-balanced Sampling}
The main idea of size-balanced sampling is to repeat the data sampling for the COVID-19 cases with small infections and also the CAP cases with large infections in each mini-batch during training.
Normally, we use the uniform sampling in the entire dataset for the network training (i.e., ``Attention RN34 + US" branch in Fig. \ref{fig:framework}). Specifically, each sample in the training dataset is fed into the network only once with equal probability within one epoch. Thus, the model can review the entire dataset when maintaining the intrinsic data distribution. Due to the imbalance of the distribution of infection size, we train a second network via the size-balanced sampling strategy (i.e., ``Attention RN34 + SS" branch). It aims to boost the sampling possibility of the small-infection-area COVID-19 and also large-infection-area CAP cases in each mini-batch. To this end, we split the data into 4 groups according to the volume ratio of the pneumonia infection regions and the lung: 1) small-infection-area COVID-19, 2) large-infection-area COVID-19, 3) small-infection-area CAP, and 4) large-infection-area CAP. For COVID-19, we define the cases that meet the criteria of $<0.030$ as small-infection-area COVID-19, and the rest as large-infection-area COVID-19. For CAP, we define the cases with the ratio $>0.001$ as large-infection-area CAP and the rest as small-infection-area CAP. We define the numbers of samples for the 4 groups as $[N_{small}^{covid},N_{large}^{covid},N_{small}^{cap},N_{large}^{cap}]$. Then, inspired by the class-resampling strategy in \cite{zhou2019bbn}, we define the weights $[W_{small}^{covid},W_{large}^{covid},W_{small}^{cap},W_{large}^{cap}]$ for 4 groups as $[N_{large}^{covid}/N_{small}^{covid},1,1,N_{small}^{cap}/N_{large}^{cap}]$. Since the numbers of small-infection-area COVID-19 and large-infection-area CAP are relatively small, the weights $W_{small}^{covid}$ and $W_{large}^{cap}$ are higher than 1. The values of these two weights are approximately 1.5 in each training fold. Then, the sampling possibilities for 4 groups are calculated by the weight of each group divided by the sum of all weights, $W_{sum}$. 
In a mini-batch, we randomly select a group according to the refined possibilities for each group $[W_{small}^{covid}/W_{sum},1/W_{sum},1/W_{sum},W_{large}^{cap}/W_{sum}]$, and uniformly pick up a sample from the selected group. This strategy ensures to have more possibility to sample cases from the two groups of 1) COVID-19 with small infections and 2) CAP with large infections. We conduct the size-balanced sampling strategy for all mini-batches when training the ``Attention RN34 + SS" model.

\subsection{Objective Function}
Two losses are used to train ``Attention RN34 + US" and ``Attention RN34 + SS" models, i.e., the classification loss $L_c$ and the extra attention loss $L_{ex}$ for COVID-19 cases, respectively. We adopt the binary cross entropy as constrain for the COVID-19/CAP classification loss $L_c$. For the COVID-19 cases, given the pneumonia infection segmentation mask $M$, we can use them to directly refine the attention maps from our model and $L_{ex}$ is thus formulated as:
\begin{normalsize}
\begin{equation}
\label{eq:la}
{L_{ex}} = \frac{{\sum\nolimits_{ijk} {{{(T({A_{ijk}}) - {M_{ijk}})}^2}} }}{{\sum\nolimits_{ijk} {T({A_{ijk}}) + \sum\nolimits_{ijk} {{M_{ijk}}} } }},
\end{equation}
\end{normalsize}
where $T(A_{ijk})$ is the attention map generated from our online attention module (Eq. \ref{eq:thresholding}), and $i$, $j$ and $k$ represent the $(i,j,k)^{th}$ voxel in the attention map. The proposed $L_{ex}$ is modified from the traditional mean square error (MSE) loss, using the sum of regions of attention map $T(A_{ijk})$ and the corresponding mask $M_{ijk}$ as an adaptive normalization factor. It can adjust the loss value dynamically according to the sizes of pneumonia infection regions. Then, the overall objective function for training ``Attention RN34 + US" and ``Attention RN34 + SS" models is expressed as:
\begin{normalsize}
\begin{equation}
\label{eq:loss}
{L_{total}} = {L_c} + \lambda {L_{ex}},
\end{equation}
\end{normalsize}
where $\lambda$ is a weight factor for the attention loss. It is set to 0.5 in our experiments. For the CAP cases, only the classification loss $L_c$ is used for model training.

\subsection{Ensemble Learning}
The size-balanced sampling method could gain more attention on the minority classes and remedy the infection area bias in COVID-19 and CAP patients. A drawback is that it may suffer from the possible over-fitting of these minority classes. In contrast, the uniform sampling method could learn feature representation from the original data distribution in a relatively robust way. Taking the advantages of both sampling methods, we propose a dual-sampling method via an ensemble learning layer, which gauges the weights for the prediction results produced by the two models.

After training the two models with different sampling strategies, we use an ensemble learning layer to integrate the predictions from two models into the final diagnosis result. We combine the prediction scores with different weights for different ratios of the pneumonia infection regions and the lung:
\begin{normalsize}
\begin{equation}
\label{eq:final}
{P_{final}} = w P_{US} + (1-w) P_{SS},
\end{equation}
\end{normalsize}
where, $w$ is the weight factor. In our experiment, it is set to 0.35 for the case where the ratio meets the criterion $<0.001$ or $>0.030$, and 0.96 for the rest cases. The factor values are determined with a hyperparameter search on the TV set. Then, $P_{final}$ is the final prediction result of the dual-sampling model. As presented in Eq. \ref{eq:final}, the dual-sampling strategy combines the characteristics of uniform sampling and size-balanced sampling. For the minority classes, i.e., COVID-19 with small infections as well as CAP with large infections, we assign extra weights to the ``Attention RN34 + SS" model. For the rest cases, more weights are assigned to the ``Attention RN34 + US" model.

\section{EXPERIMENTAL RESULTS}
\label{experiment}
\subsection{Dataset}

In this study, we use a large multi-center CT data for evaluating the proposed method in diagnosis of COVID-19. In particular, we have collected a total of 4982 ($<$2mm) chest CT images from 3645 patients, including 3389 COVID-19 CT images and 1593 CAP CT images. All recruited COVID-19 patients were confirmed by RT-PCR test. 
Here, the images were provided by the Tongji Hospital of Huazhong University of Science and Technology, Shanghai Public Health Clinical Center of Fudan University, the Second Xiangya Hospital of Central South University, China-Japan Union Hospital of Jilin University, Ruijin Hospital Affiliated to Shanghai Jiao Tong University School of Medicine, Affiliated Hangzhou First People's Hospital of Zhejiang University, the Beijing Chaoyang Hospital of Capital Medical University, and Sichuan University West China Hospital. According to the data collection dates, we separate them into two datasets. The first dataset (TV dataset) is used for training and cross-validation, which includes 1094 COVID-19 images and 1092 CAP images. The second dataset serves for independent testing, including 2295 COVID-19 images and 501 CAP images. Note that the split is done on patient level, which means the images of same subject are kept in the same group of training or testing. More details are shown in Table \ref{table:dataset}.

\begin{table}[!t]
\newcommand{\tabincell}[2]{\begin{tabular}{@{}#1@{}}#2\end{tabular}}
\centering
\caption{Demographic of the training-validation (TV) dataset and test dataset. The results of ``Age" is presented as median values (range). 
}
\label{table:dataset}
\begin{tabular}{l|l|c|c}
\hline
\multicolumn{2}{l|}{Characteristics} & TV set & Test set \\
\hline
\multicolumn{4}{l}{No. (images (patients))} \\
\hline
 \multicolumn{2}{l|}{\ COVID-19} & 1094 (960) &  2295 (1605) \\
 \multicolumn{2}{l|}{\ CAP} & 1092 (628) & 501 (452) \\
 \multicolumn{2}{l|}{\ Total} & 2186 (1588) & 2796 (2057) \\
\hline
\multicolumn{4}{l}{Age (years) } \\
\hline
 \multicolumn{2}{l|}{\ COVID-19} & 50.0 (14-89) &  50.0 (8-95) \\
 \multicolumn{2}{l|}{\ CAP} & 57.0 (12-94) & 42.0 (15-98) \\
 \multicolumn{2}{l|}{\ Total} & 53.0 (12-94) &  49.0 (8-98) \\
 \hline
\multicolumn{4}{l}{Female/Male } \\
\hline
 \multicolumn{2}{l|}{\ COVID-19} & 479/481 & 800/805 \\
 \multicolumn{2}{l|}{\ CAP} & 322/306 & 255/197 \\
 \multicolumn{2}{l|}{\ Total} & 801/787 & 1055/1002 \\
\hline
\end{tabular}
\end{table}

Thin-slice chest CT images are used in this study with the CT thickness ranging from 0.625 to 1.5mm.  CT scanners include uCT 780 from UIH, Optima CT520, Discovery CT750, LightSpeed 16 from GE, Aquilion ONE from Toshiba, SOMATOM Force from Siemens, and SCENARIA from Hitachi. Scanning protocol includes: 120 kV, with breath hold at full inspiration. All CT images are anonymized before sending them for conducting this research project. The study is approved by the Institutional Review Board of participating institutes. Written informed consent is waived due to the retrospective nature of the study. 

\subsection{Image pre-processing}
Data are pre-processed in the following steps before feeding them into the network. First, we resample all CT images and the corresponding masks of lungs and infection regions to the same spacing (0.7168mm, 0.7168mm, 1.25mm for the x, y, and z axes, respectively) for the normalization to the same voxel size. Second, we down-sample the CT images and segmentation masks into the approximately half sizes considering efficient computation. To avoid morphological change in down-sampling, we use the same scale factor in all three dimensions and pad zeros to ensure the final size of $138\times256\times256$. We should emphasize that our method is capable of handling full-size images. Third, we conduct ``window/level" (window: 1500, level: -600) scaling in CT images for contrast enhancement. We truncate the CT image into the window [-1350, 150], which sets the intensity value above 150 to 150, and below -1350 to -1350. Finally, following the standard protocol of data pre-processing, we normalize the voxel-wise intensities in the CT images to the interval $[0,1]$. 

\subsection{Training Details and Evaluation Methods}
We implement the networks in PyTorch \cite{paszke2019pytorch}, and use NVIDIA Apex for less memory consumption and faster computation. We also use the Adam \cite{kingma2014adam} optimizer with momentum set to 0.9, a weight decay of 0.0001, and a learning rate of 0.0002 that is reduced by a factor of 10 after every 5 epochs. We set the batch size as 20 during the training. In our experiments, all the models are trained from scratch. In the TV set, we conduct 5-fold cross-validation. In each fold, the model is evaluated on the validation set in the end of each training epoch. The best checkpoint model with the best evaluation performance within 20 epochs is used as the final model and then evaluated on the test set. All the models are trained in 4 NVIDIA TITAN RTX graphics processing units, and the inference time for one sample is approximately 4.6s in one NVIDIA TITAN RTX GPU. For evaluating, we use five different metrics to measure the classification results from the model: area under the receiver operating characteristic curve (AUC), accuracy, sensitivity, specificity, and F1-score. AUC represents degree or measure of separability. In this study, we calculated the accuracy, sensitivity, specificity, and F1-score at the threshold of 0.5.

\subsection{Results}

\begin{table}[!t]
\newcommand{\tabincell}[2]{\begin{tabular}{@{}#1@{}}#2\end{tabular}}
\caption{Comparasion of classification results of differnet models on the TV set and test set (RN34: 3D ResNet34; US: Uniform Sampling; SS: Size-balanced Sampling; DS: Dual-sampling). The results of AUC, accuracy, sensitivity, specificity and F1-score are present in this table. The results on TV set are the combined results of 5 validation sets. For results on the test set, we show mean$\pm$std (standard deviation) scores of five trained models of each training-validation fold.}
\label{table:results}
\centering
\begin{tabular}{l|l|c|c}
\hline
\multicolumn{2}{l|}{Results} & \tabincell{c}{TV set} & Test set  \\
\hline
\multirow{4}{*}{AUC} & RN34 + US & 0.984 & 0.934$\pm$0.011 \\
 & Attention RN34 + US & 0.986 & \textbf{0.948$\pm$0.003} \\
 & Attention RN34 + SS & 0.987 & 0.938$\pm$0.002 \\
 & Attention RN34 + DS & \textbf{0.988} & 0.944$\pm$0.003 \\
 \hline
 \multirow{4}{*}{Accuracy} & RN34 + US & 0.945 & 0.859$\pm$0.013 \\
 & Attention RN34 + US & 0.947 & \textbf{0.879$\pm$0.012} \\
 & Attention RN34 + SS & 0.951 & 0.869$\pm$0.008 \\
 & Attention RN34 + DS & \textbf{0.954} & 0.875$\pm$0.009 \\
 \hline
 \multirow{4}{*}{Sensitivity} & RN34 + US & 0.931 & 0.856$\pm$0.029 \\
 & Attention RN34 + US & 0.941 & \textbf{0.872$\pm$0.018} \\
 & Attention RN34 + SS & 0.953 & 0.868$\pm$0.020 \\
 & Attention RN34 + DS & \textbf{0.954} & 0.869$\pm$0.016 \\
 \hline
 \multirow{4}{*}{Specificity} & RN34 + US & \textbf{0.959} & 0.870$\pm$0.071 \\
 & Attention RN34 + US & 0.953 & \textbf{0.907$\pm$0.029} \\
 & Attention RN34 + SS & 0.948 & 0.876$\pm$0.048 \\
 & Attention RN34 + DS & 0.954 & 0.901$\pm$0.025 \\
 \hline
 \multirow{4}{*}{F1-score} & RN34 + US & 0.945 & 0.798$\pm$0.011 \\
 & Attention RN34 + US & 0.947 & \textbf{0.825$\pm$0.013} \\
 & Attention RN34 + SS & 0.951 & 0.811$\pm$0.004 \\
 & Attention RN34 + DS & \textbf{0.954} & 0.820$\pm$0.008 \\
 \hline
\end{tabular}
\end{table}

\begin{figure*}[!t]
\centering
\includegraphics[width=14cm, height=10cm]{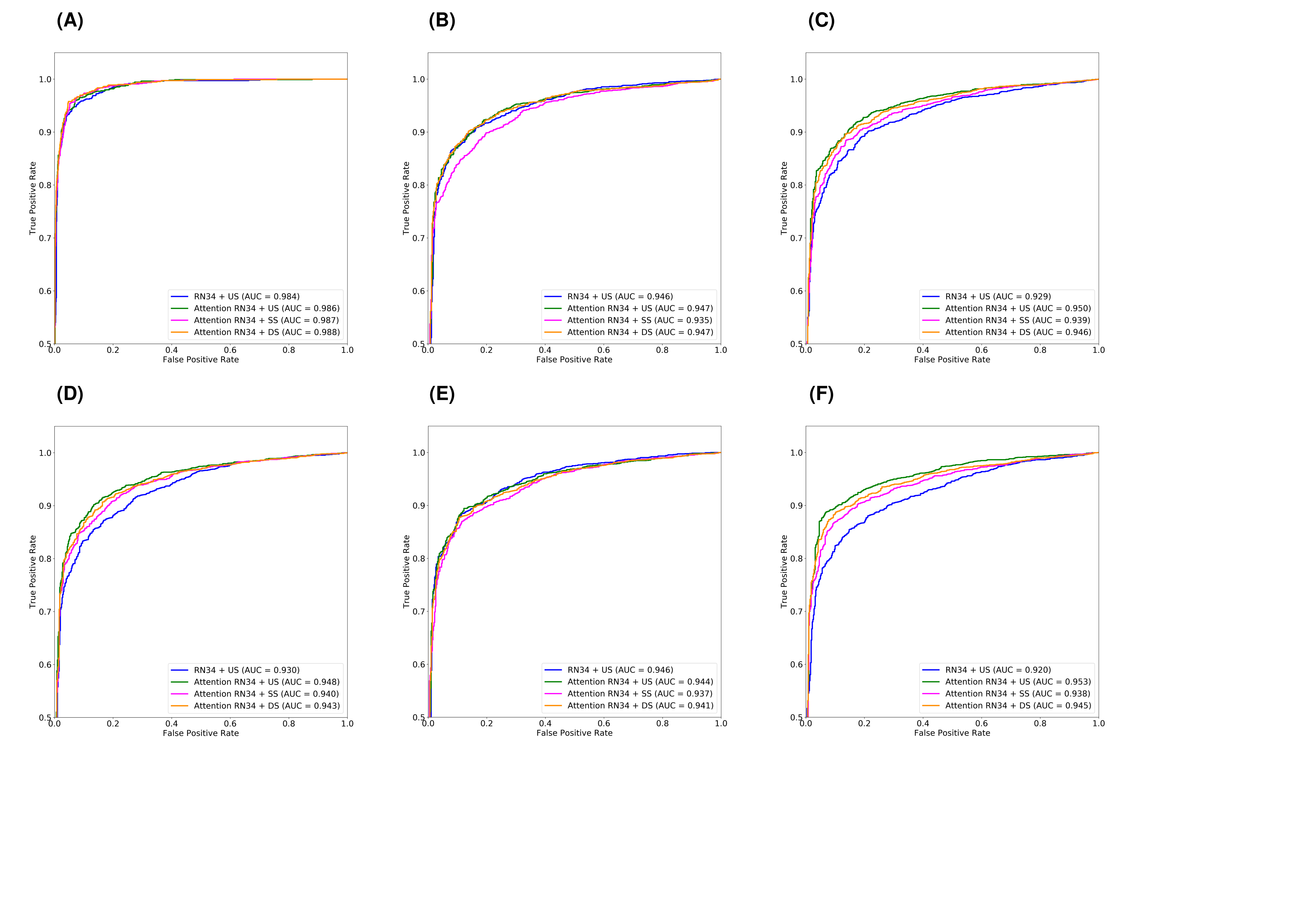}
\caption{ROC curves of the TV set and the test set. (A) ROC curves of TV set for 5 folds. (B) ROC curve of test set by using the model from TV set fold 1. (C) ROC curve of test set by using the model from TV set fold 2. (D) ROC curve of test set by using the model from TV set fold 3. (E) ROC curve of test set by using the model from TV set fold 4. (F) ROC curve of test set by using the model from TV set fold 5.} 
\label{fig:roc-train}
\end{figure*}

First, we conduct 5-fold cross-validation on the TV set. The experimental results are shown in Table \ref{table:results}, which combines the results of all 5 validation sets. The receiver operating characteristic (ROC) curve is also shown in Fig. \ref{fig:roc-train}(A). We can see that the models with the proposed attention refinement technique can improve the AUC and sensitivity scores.  
At the same time, we can see that ``Attention RN34 + DS" achieves the highest performance in AUC, accuracy, sensitivity, and F1-score, when combining the two models with different sampling strategies. As for the specificity, the performance of the dual-sampling method is a little bit lower than that of ResNet34 with uniform sampling. 

We further investigate the generalization capability of the model by deploying the five trained models of five individual folds on the independent testing dataset. From Fig. \ref{fig:roc-train}(B-F), we can see that the trained model of each fold achieves similar performance, implying consistent performance with different training data. Compared with the results on the TV set in Fig. 4(A), the AUC score of the models with the proposed attention module (``Attention RN34 + DS") on the independent test set drops from 0.988 to 0.944, while the AUC score of ``RN34 + US" drops from 0.984 to 0.934. This indicates the strong robustness of our model, trained with our attention module, against possible over-fitting. The proposed attention module can also ensure that the decisions made by the model depend mainly on the infection regions, suppressing the contributions from the non-related parts in the images. All 501 CAP images in the test set are from a single site that was not included in the TV set. ``Attention RN34 + US" and ``Attention RN34 + DS" models achieves $\geq 90.0\%$ in specificity for these images.
We can see that our algorithm maintains a great performance on the data acquired from different centers. In the next section, the effects of different sampling strategies are presented. In order to confirm whether there exists significant difference when using the proposed attention module or not, paired $t$-tests are applied. The $p$-values between ``RN34 + US" and the three proposed methods are calculated. All the $p$-values are small than 0.01, implying that the proposed methods have significant improvements compared with ``RN34 + US".

\begin{table*}[!t]
\newcommand{\tabincell}[2]{\begin{tabular}{@{}#1@{}}#2\end{tabular}}
\caption{Group-wise results on TV set and test set. Based on the volume ratio of pneumonia regions and the lung, the data is divided into 3 groups: the volume ratios that meet the criteria of $<0.005$, $0.005-0.030$, and $>0.030$, respectively.}
\label{table:group}
\centering
\begin{tabular}{l|l|c|c|c|c|c|c}
\hline
\multicolumn{2}{l|}{\multirow{2}{*}{Results}} & \multicolumn{3}{c|}{TV set} & \multicolumn{3}{c}{Test set} \\
\cline{3-8}
 \multicolumn{2}{c|}{} & $<0.005$ & $0.005-0.030$ & $>0.030$ & $<0.005$ & $0.005-0.030$ & $>0.030$  \\
 \hline
\multirow{3}{*}{\tabincell{l}{No. of \\ images}} & COVID-19 & 151 & 318 & 625 & 363 & 718 & 1214 \\
  & \tabincell{l}{CAP} & 838 & 183 & 71 & 436 & 41 & 24 \\
  & Total No. & 989 & 501 & 696 & 799 & 759 &  1238 \\
 \hline
\multirow{4}{*}{AUC} & RN34 + US & 0.949 & 0.975 & 0.972 &  0.796$\pm$0.032 &0.914$\pm$0.021 &0.905$\pm$0.011 \\
 & Attention RN34 + US & 0.958 & 0.974 & 0.986 &  \textbf{0.835$\pm$0.012} & \textbf{0.923$\pm$0.005} &0.906$\pm$0.016 \\
 & Attention RN34 + SS & 0.958 & \textbf{0.981} & 0.986 &  0.816$\pm$0.007 &0.919$\pm$0.004 & 0.906$\pm$0.014   \\
 & Attention RN34 + DS & \textbf{0.960} & \textbf{0.981} & \textbf{0.987} &  0.830$\pm$0.011&0.919$\pm$0.004& \textbf{0.907$\pm$0.015}  \\
 \hline
 \multirow{4}{*}{Accuracy} & RN34 + US & 0.930 & 0.930 & 0.976 &  0.719$\pm$0.015 &0.848$\pm$0.037 &0.955$\pm$0.007  \\
 & Attention RN34 + US & 0.932 & 0.930 & \textbf{0.981} &  0.752$\pm$0.017 & \textbf{0.871$\pm$0.017} & \textbf{0.965$\pm$0.008} \\
 & Attention RN34 + SS & 0.938 & \textbf{0.942} & 0.974 &  0.747$\pm$0.006 &0.858$\pm$0.018 &0.955$\pm$0.009   \\
 & Attention RN34 + DS & \textbf{0.941} & \textbf{0.942} & \textbf{0.981} &  \textbf{0.755$\pm$0.012} &0.859$\pm$0.016&0.962$\pm$0.007  \\
 \hline
 \multirow{4}{*}{Sensitivity} & RN34 + US & 0.675 & 0.925 & 0.995 &  0.514$\pm$0.093 &0.851$\pm$0.042 &0.962$\pm$0.007 \\
 & Attention RN34 + US & 0.722 & 0.937 & \textbf{0.997} &  0.534$\pm$0.050 & \textbf{0.875$\pm$0.021} & \textbf{0.972$\pm$0.008} \\
 & Attention RN34 + SS & \textbf{0.815} & \textbf{0.953} & 0.987 &  \textbf{0.569$\pm$0.061} &0.862$\pm$0.020 &0.960$\pm$0.010   \\
 & Attention RN34 + DS & 0.795 & \textbf{0.953} & 0.994 &  0.549$\pm$0.049&0.863$\pm$0.018&0.968$\pm$0.008  \\
 \hline
 \multirow{4}{*}{Specificity} & RN34 + US & \textbf{0.976} & \textbf{0.940} & 0.803 &  0.889$\pm$0.074 & \textbf{0.810$\pm$0.078} &0.617$\pm$0.062  \\
 & Attention RN34 + US & 0.970 & 0.918 & 0.845 &  \textbf{0.933$\pm$0.024} &0.785$\pm$0.090 &0.642$\pm$0.037 \\
 & Attention RN34 + SS & 0.961 & 0.923 & 0.859 &  0.896$\pm$0.051 &0.785$\pm$0.047 & \textbf{0.667$\pm$0.051}   \\
 & Attention RN34 + DS & 0.968 & 0.923 & \textbf{0.873} &  0.926$\pm$0.025& 0.790$\pm$0.051 &0.650$\pm$0.037 \\
 \hline
 \multirow{4}{*}{F1-score} & RN34 + US & 0.853 & 0.926 & 0.928 &  0.698$\pm$0.022 &0.643$\pm$0.035 &0.663$\pm$0.018  \\
 & Attention RN34 + US & 0.863 & 0.925 & 0.946 &  0.732$\pm$0.022& \textbf{0.662$\pm$0.015} & \textbf{0.702$\pm$0.026} \\
 & Attention RN34 + SS & 0.882 & \textbf{0.938} & 0.929 &  0.732$\pm$0.009&0.648$\pm$0.020&0.671$\pm$0.018  \\
 & Attention RN34 + DS & \textbf{0.885} & \textbf{0.938} & \textbf{0.947} &  \textbf{0.737$\pm$0.017} &0.649$\pm$0.017&0.692$\pm$0.018  \\
 \hline
\end{tabular}
\end{table*}

\begin{figure*}[!t]
\centering
\includegraphics[width=14.6cm, height=12cm]{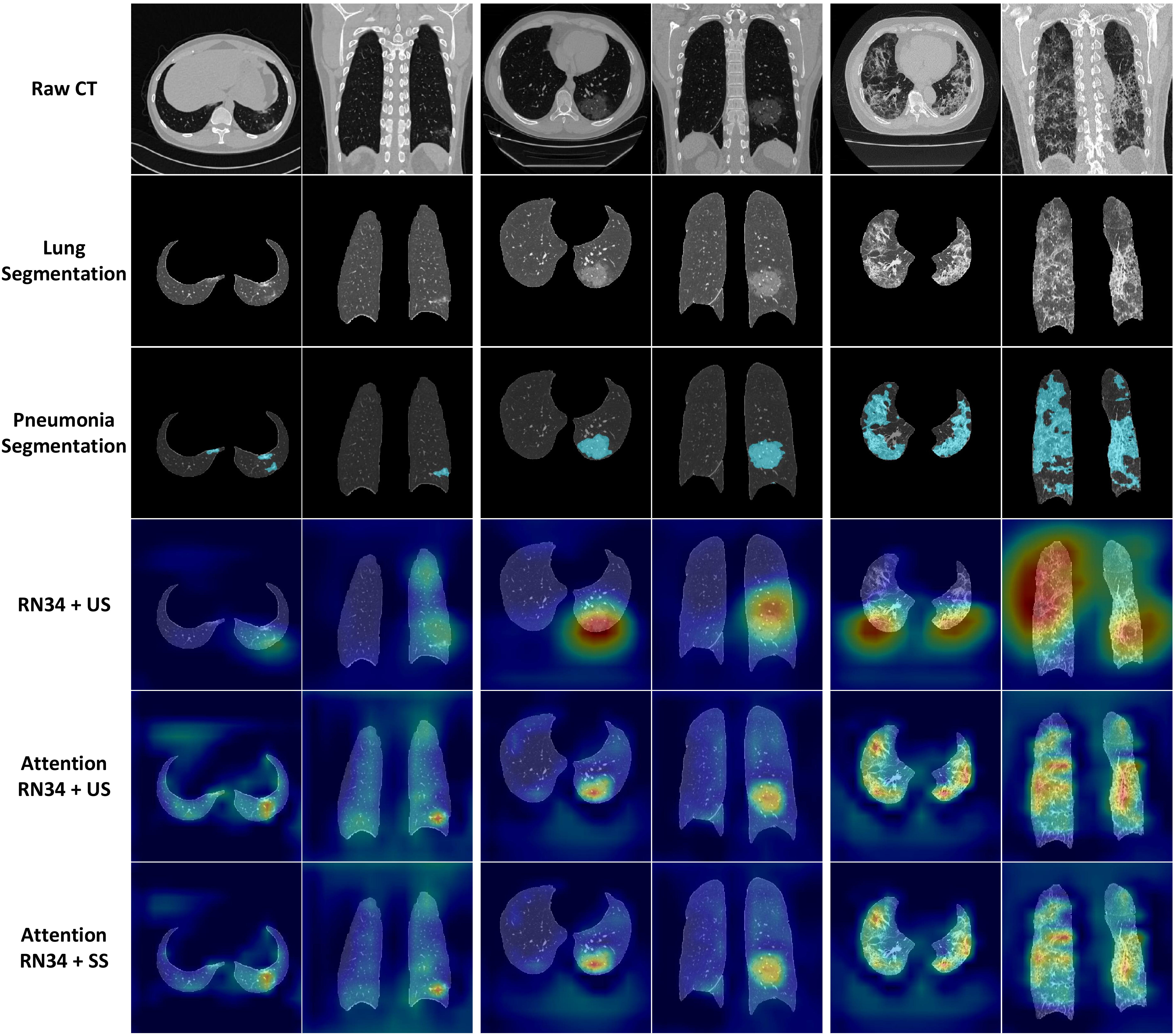}
\caption{Visualization results of our methods on three COVID-19 cases from small-infection group ($<0.005$), median-infection group ($0.005-0.030$) and large-infection group ($>0.030$) of the test set are shown from left to right, respectively. For each case, we show the visualization results in both axial view and coronal view. We show the original images (first row), and the segmentation results of the lung and pneumonia infection regions (2$^{nd}$ and 3$^{rd}$ rows) by the VB-Net tookit \cite{shan+2020lung}. For the attention results, we show the Grad-CAM results of ``RN34 +US" (4$^{th}$ row), and the attention maps obtained by our proposed attention module of ``Attention RN34 + US" and ``Attention RN34 + SS" models (5$^{th}$ and 6$^{th}$ rows).} 
\label{fig:visualization}
\end{figure*}

\subsection{Detailed Analysis}
To demonstrate the effectiveness in diagnosing pneumonia of different severity, we use the VB-Net toolkit \cite{shan+2020lung} to get the lung mask and the pneumonia infection regions for all CT images. Based on the quantified volume ratio of pneumonia infection regions over the lung, we roughly divide the data into 3 groups in both the TV set and the test set, according to the ratios, i.e., 1) $<0.005$, 2) $0.005-0.030$, and 3) $>0.030$. As shown in Table \ref{table:group}, most of COVID-19 images have high ratios (higher than 0.030), while most CAPs are lower than 0.005, which may indicate that the severity of COVID-19 is usually higher than that of CAP in our collected dataset. Furthermore, the classification results of COVID-19 is highly related with the ratio. In Table \ref{table:group}, we can see that the sensitivity scores are relatively high for the high infected region group ($>0.030$), while the specificity scores are relatively low for the small infection region group ($<0.005$). This performance matches the nature of COVID-19 and CAP in the collected dataset.

As size-balanced sampling strategy (``Attention RN34 + SS") is applied in the training procedure, we can find that the sensitivity of the small infected region group ($<0.005$) increases from 0.534 to 0.569, compared with the case of using the uniform sampling strategy (``Attention RN34 + US"). And also the specificity of the large infected region group ($>0.030$) increases from 0.642 to 0.667. These results demonstrate that the size-balanced sampling strategy can effectively improve the classification robustness when the bias of the pneumonia area exists. However, if we only utilize the size-balanced sampling strategy in the training process, the sensitivity of the large infected region group ($>0.030$) will decrease from 0.965 to 0.955, and the specificity of the small infected region group ($<0.005$) will decrease from 0.933 to 0.896. This reflects that some advantages of the network may be sacrificed in order to achieve specific requirements. To achieve a dynamic balance between the two extreme conditions, we present the results using the ensemble learning with the dual-sampling model (i.e., ``Attention RN34 + DS"). From the sensitivity and specificity in both small and large infected region groups, dual sampling strategy can preserve the classification ability obtained by uniform sampling, and slightly improve the classification performance of the COVID-19 cases in the small infected region group and the CAP cases in the large infected region group. Furthermore, the $p$-values between ``Attention RN34 + US" and ``Attention RN34 + DS" in both small-infected-region group ($<0.005$) and high-infected-region group ($>0.030$) are calculated. All the $p$-values are smaller than 0.01, which also proves the effectiveness and necessity of the dual sampling strategy.


Finally, we show typical attention maps obtained by our models (Fig. \ref{fig:visualization}) trained in one fold. For comparison, we show the attention results of naive ReNset34 (``RN34 + US") in the same fold without both the online attention module and the infection mask refinement, and perform the model explanation techniques (Grad-CAM \cite{selvaraju2017grad}) to get the heatmaps for classification. We can see that the output of Grad-CAM roughly indicates the infection localization, yet sometimes appears far outside of the lung. However, the attention maps from our models (``Attention RN34 + US" and ``Attention RN34 + SS") can reveal the precise locations of the infection.
These conspicuous areas in attention maps are similar to the infection segmentation results, which demonstrates that the final classification results determined by our model are reliable and interpretable. The attention maps thus can be possibly used as the basis to derive the COVID-19 diagnosis in clinical practice.


\begin{figure}[!t]
\centering
\includegraphics[width=8.6cm, height=10cm]{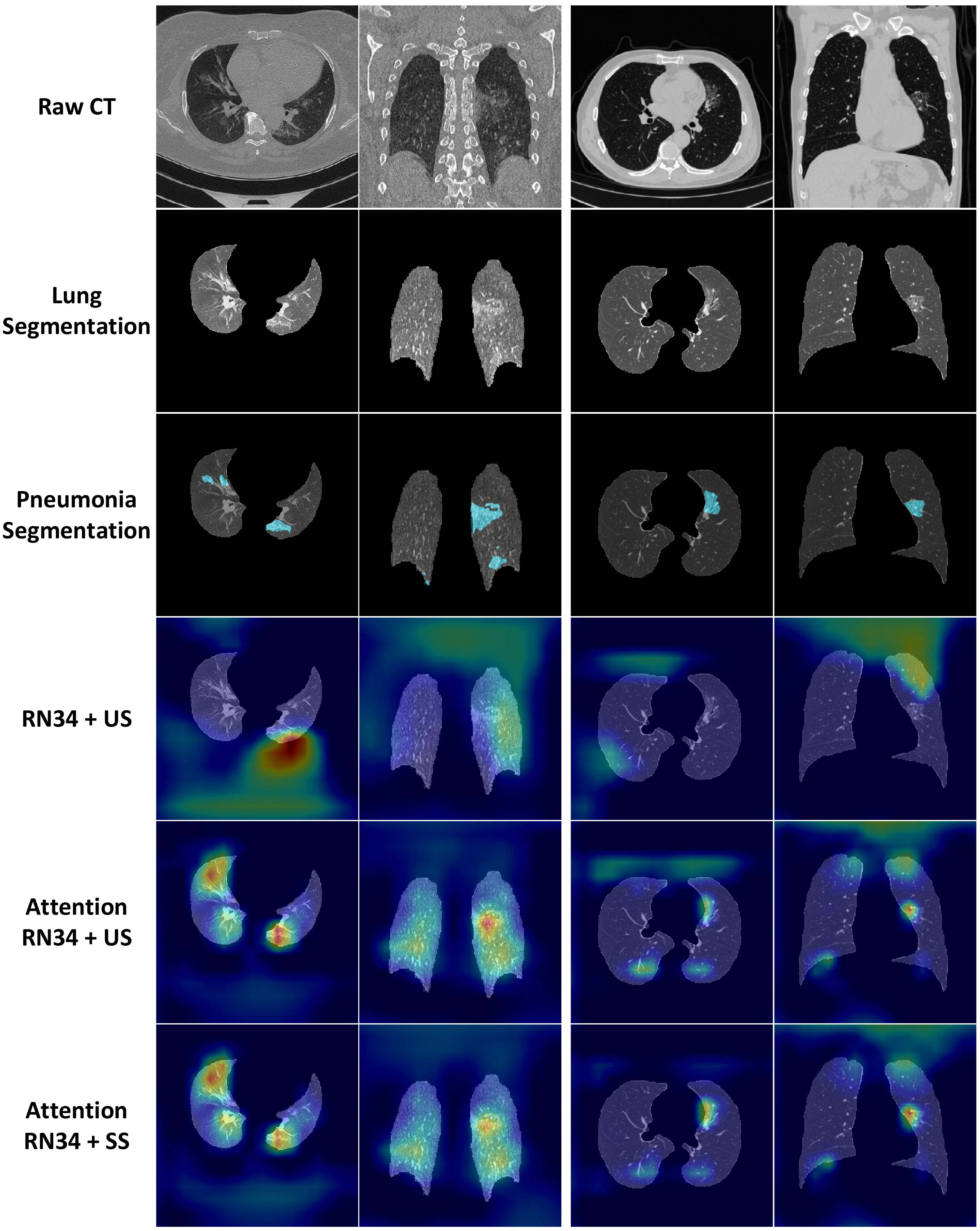}
\caption{Visualization results of two failure cases.}
\label{fig:visualization_failure}
\end{figure}

\subsection{Failure Analysis}
We also show two failure cases in Fig. \ref{fig:visualization_failure}, where the COVID-19 cases are classified as CAP by mistake for all the models. As can be observed from the results shown in Fig. \ref{fig:visualization}, the attention maps from all the models incorrectly get activated on many areas unrelated to pneumonia. ``RN34 + US" model even generates many highlighted areas in the none-lung region instead of focusing on lungs. With the proposed attention constrain, the attention maps of ``Attention RN34 + US" and ``Attention RN34 + SS" have partially alleviated this problem. But still the visual evidences are insufficient to reach a final correct prediction.

\section{DISCUSSION AND CONCLUSION}
For COVID-19, it is important to get the diagnosis result at soon as possible. Although RT-PCR is the current ground truth to diagnose COVID-19, it will take up to days to get the final results and the capacity of the tests is also limited in many places especially in the early outbreak \cite{ai2020correlation}. CT is shown as a powerful tool and could provide the chest scan results in several minutes. It is beneficial to develop an automatic diagnosis method based on chest CT to assist the COVID-19 screening.
In this study, we explore a deep-learning-based method to perform automatic COVID-19 diagnosis from CAP in chest CT images. We evaluate our method by the largest multi-center CT data in the world, to the best of our knowledge. To further evaluate the generalization ability of the model, we use independent data from different hospitals (not included in the TV set), achieving AUC of 0.944, accuracy of 87.5\%, sensitivity of 86.9\%, specificity of 90.1\%, and F1-score of 82.0\%. At the same time, to better understand the decision of the deep learning model, we also refine the attention module and show the visual evidence, which is able to reveal important regions used in the model for diagnosis. Our proposed method could be further extended for differential diagnosis of pneumonia, which can greatly assist physicians.

There also exist several limitations in this study. First, 
when longitudinal data becomes ready, the proposed model should be tested for its consistency tracking the development of the COVID-19 during the treatment, as considered in \cite{xue2006classic}. Second, although the proposed online attention module could largely improve the interpretability and explainability in COVID-19 diagnosis, in comparison to the conventional methods such as Grad-CAM, future work is still needed to analyze the correlation between these attention localizations with the specific imaging signs that are frequently used in clinical diagnosis. There also exist some failure cases that the visualization results do not appear correctly at the pneumonia infection regions, as shown in Fig. \ref{fig:visualization_failure}. This motivates us to further improve the attention module to better focus on the related regions and reduce the distortion from cofounding visual information to the classification task in the future research.
Third, we also notice that the accuracy of the small-infection-area COVID-19 is not quite satisfactory. This indicates the necessity of combining CT images with clinical assessment and laboratory tests for precise diagnosis of early COVID-19, which will also be covered by our future work. The last but not least, the CAP cases used in this study do not include the subtype information, i.e., bacterial, fungal, and non-COVID-19 viral pneumonia. To assist the clinical diagnosis of pneumonia subtypes would also be beneficial.

To conclude, we have developed a 3D CNN network with both online attention refinement and dual-sampling strategy to distinguish COVID-19 from the CAP in the chest CT images. The generalization performance of this algorithm is also verified by the largest multi-center CT data in the world, to our best knowledge.
\ifCLASSOPTIONcaptionsoff
  \newpage
\fi



%



\normalem
\bibliography{sample.bib}

\begin{thebibliography}{10}
\providecommand{\url}[1]{#1}
\csname url@samestyle\endcsname
\providecommand{\newblock}{\relax}
\providecommand{\bibinfo}[2]{#2}
\providecommand{\BIBentrySTDinterwordspacing}{\spaceskip=0pt\relax}
\providecommand{\BIBentryALTinterwordstretchfactor}{4}
\providecommand{\BIBentryALTinterwordspacing}{\spaceskip=\fontdimen2\font plus
\BIBentryALTinterwordstretchfactor\fontdimen3\font minus
  \fontdimen4\font\relax}
\providecommand{\BIBforeignlanguage}[2]{{%
\expandafter\ifx\csname l@#1\endcsname\relax
\typeout{** WARNING: IEEEtran.bst: No hyphenation pattern has been}%
\typeout{** loaded for the language `#1'. Using the pattern for}%
\typeout{** the default language instead.}%
\else
\language=\csname l@#1\endcsname
\fi
#2}}
\providecommand{\BIBdecl}{\relax}
\BIBdecl

\bibitem{world2020coronavirus80}
WHO, ``Coronavirus disease 2019 (covid-19): situation report, 80,'' 2020.

\bibitem{world2020director}
------, ``Who director-general's remarks at the media briefing on 2019-ncov on
  11 february 2020. 2020,'' 2020.

\bibitem{world2020coronavirus}
------, ``Coronavirus disease (covid-2019) situation reports,'' 2020.

\bibitem{wu2020characteristics}
Z.~Wu and J.~M. McGoogan, ``Characteristics of and important lessons from the
  coronavirus disease 2019 (covid-19) outbreak in china: summary of a report of
  72 314 cases from the chinese center for disease control and prevention,''
  \emph{Jama}, 2020.

\bibitem{mahase2020coronavirus}
E.~Mahase, ``Coronavirus: covid-19 has killed more people than sars and mers
  combined, despite lower case fatality rate,'' 2020.

\bibitem{zu2020coronavirus}
Z.~Y. Zu, M.~D. Jiang, P.~P. Xu, W.~Chen, Q.~Q. Ni, G.~M. Lu, and L.~J. Zhang,
  ``Coronavirus disease 2019 (covid-19): A perspective from china,''
  \emph{Radiology}, p. 200490, 2020.

\bibitem{chan2020familial}
J.~F.-W. Chan, S.~Yuan, K.-H. Kok, K.~K.-W. To, H.~Chu, J.~Yang, F.~Xing,
  J.~Liu, C.~C.-Y. Yip, R.~W.-S. Poon \emph{et~al.}, ``A familial cluster of
  pneumonia associated with the 2019 novel coronavirus indicating
  person-to-person transmission: a study of a family cluster,'' \emph{The
  Lancet}, vol. 395, no. 10223, pp. 514--523, 2020.

\bibitem{ai2020correlation}
T.~Ai, Z.~Yang, H.~Hou, C.~Zhan, C.~Chen, W.~Lv, Q.~Tao, Z.~Sun, and L.~Xia,
  ``Correlation of chest ct and rt-pcr testing in coronavirus disease 2019
  (covid-19) in china: a report of 1014 cases,'' \emph{Radiology}, p. 200642,
  2020.

\bibitem{chung2020ct}
M.~Chung, A.~Bernheim, X.~Mei, N.~Zhang, M.~Huang, X.~Zeng, J.~Cui, W.~Xu,
  Y.~Yang, Z.~A. Fayad \emph{et~al.}, ``Ct imaging features of 2019 novel
  coronavirus (2019-ncov),'' \emph{Radiology}, p. 200230, 2020.

\bibitem{shan+2020lung}
F.~Shan, Y.~Gao, J.~Wang, W.~Shi, N.~Shi, M.~Han, Z.~Xue, D.~Shen, and Y.~Shi,
  ``Lung infection quantification of covid-19 in ct images with deep
  learning,'' \emph{arXiv preprint arXiv:2003.04655}, 2020.

\bibitem{lecun2015deep}
Y.~LeCun, Y.~Bengio, and G.~Hinton, ``Deep learning,'' \emph{nature}, vol. 521,
  no. 7553, pp. 436--444, 2015.

\bibitem{krizhevsky2012imagenet}
A.~Krizhevsky, I.~Sutskever, and G.~E. Hinton, ``Imagenet classification with
  deep convolutional neural networks,'' in \emph{Advances in neural information
  processing systems}, 2012, pp. 1097--1105.

\bibitem{he2016deep}
K.~He, X.~Zhang, S.~Ren, and J.~Sun, ``Deep residual learning for image
  recognition,'' in \emph{Proceedings of the IEEE conference on computer vision
  and pattern recognition}, 2016, pp. 770--778.

\bibitem{huang2017densely}
G.~Huang, Z.~Liu, L.~Van Der~Maaten, and K.~Q. Weinberger, ``Densely connected
  convolutional networks,'' in \emph{Proceedings of the IEEE conference on
  computer vision and pattern recognition}, 2017, pp. 4700--4708.

\bibitem{nie2016estimating}
D.~Nie, X.~Cao, Y.~Gao, L.~Wang, and D.~Shen, ``Estimating ct image from mri
  data using 3d fully convolutional networks,'' in \emph{Deep Learning and Data
  Labeling for Medical Applications}.\hskip 1em plus 0.5em minus 0.4em\relax
  Springer, 2016, pp. 170--178.

\bibitem{wang2017chestx}
X.~Wang, Y.~Peng, L.~Lu, Z.~Lu, M.~Bagheri, and R.~M. Summers, ``Chestx-ray8:
  Hospital-scale chest x-ray database and benchmarks on weakly-supervised
  classification and localization of common thorax diseases,'' in
  \emph{Proceedings of the IEEE conference on computer vision and pattern
  recognition}, 2017, pp. 2097--2106.

\bibitem{ronneberger2015u}
O.~Ronneberger, P.~Fischer, and T.~Brox, ``U-net: Convolutional networks for
  biomedical image segmentation,'' in \emph{International Conference on Medical
  image computing and computer-assisted intervention}.\hskip 1em plus 0.5em
  minus 0.4em\relax Springer, 2015, pp. 234--241.

\bibitem{lecun1989backpropagation}
Y.~LeCun, B.~Boser, J.~S. Denker, D.~Henderson, R.~E. Howard, W.~Hubbard, and
  L.~D. Jackel, ``Backpropagation applied to handwritten zip code
  recognition,'' \emph{Neural computation}, vol.~1, no.~4, pp. 541--551, 1989.

\bibitem{pang2019automatic}
T.~Pang, S.~Guo, X.~Zhang, and L.~Zhao, ``Automatic lung segmentation based on
  texture and deep features of hrct images with interstitial lung disease,''
  \emph{BioMed Research International}, vol. 2019, 2019.

\bibitem{park2019lung}
B.~Park, H.~Park, S.~M. Lee, J.~B. Seo, and N.~Kim, ``Lung segmentation on hrct
  and volumetric ct for diffuse interstitial lung disease using deep
  convolutional neural networks,'' \emph{Journal of Digital Imaging}, vol.~32,
  no.~6, pp. 1019--1026, 2019.

\bibitem{yasaka2018deep}
K.~Yasaka, H.~Akai, O.~Abe, and S.~Kiryu, ``Deep learning with convolutional
  neural network for differentiation of liver masses at dynamic
  contrast-enhanced ct: a preliminary study,'' \emph{Radiology}, vol. 286,
  no.~3, pp. 887--896, 2018.

\bibitem{huang2018added}
P.~Huang, S.~Park, R.~Yan, J.~Lee, L.~C. Chu, C.~T. Lin, A.~Hussien,
  J.~Rathmell, B.~Thomas, C.~Chen \emph{et~al.}, ``Added value of
  computer-aided ct image features for early lung cancer diagnosis with small
  pulmonary nodules: a matched case-control study,'' \emph{Radiology}, vol.
  286, no.~1, pp. 286--295, 2018.

\bibitem{ardila2019end}
D.~Ardila, A.~P. Kiraly, S.~Bharadwaj, B.~Choi, J.~J. Reicher, L.~Peng, D.~Tse,
  M.~Etemadi, W.~Ye, G.~Corrado \emph{et~al.}, ``End-to-end lung cancer
  screening with three-dimensional deep learning on low-dose chest computed
  tomography,'' \emph{Nature medicine}, vol.~25, no.~6, pp. 954--961, 2019.

\bibitem{lakhani2017deep}
P.~Lakhani and B.~Sundaram, ``Deep learning at chest radiography: automated
  classification of pulmonary tuberculosis by using convolutional neural
  networks,'' \emph{Radiology}, vol. 284, no.~2, pp. 574--582, 2017.

\bibitem{irvin2019chexpert}
J.~Irvin, P.~Rajpurkar, M.~Ko, Y.~Yu, S.~Ciurea-Ilcus, C.~Chute, H.~Marklund,
  B.~Haghgoo, R.~Ball, K.~Shpanskaya \emph{et~al.}, ``Chexpert: A large chest
  radiograph dataset with uncertainty labels and expert comparison,'' in
  \emph{Proceedings of the AAAI Conference on Artificial Intelligence},
  vol.~33, 2019, pp. 590--597.

\bibitem{cruz2013deep}
A.~A. Cruz-Roa, J.~E.~A. Ovalle, A.~Madabhushi, and F.~A.~G. Osorio, ``A deep
  learning architecture for image representation, visual interpretability and
  automated basal-cell carcinoma cancer detection,'' in \emph{International
  Conference on Medical Image Computing and Computer-Assisted
  Intervention}.\hskip 1em plus 0.5em minus 0.4em\relax Springer, 2013, pp.
  403--410.

\bibitem{zhang2018visual}
Q.-s. Zhang and S.-C. Zhu, ``Visual interpretability for deep learning: a
  survey,'' \emph{Frontiers of Information Technology \& Electronic
  Engineering}, vol.~19, no.~1, pp. 27--39, 2018.

\bibitem{zhou2016learning}
B.~Zhou, A.~Khosla, A.~Lapedriza, A.~Oliva, and A.~Torralba, ``Learning deep
  features for discriminative localization,'' in \emph{Proceedings of the IEEE
  conference on computer vision and pattern recognition}, 2016, pp. 2921--2929.

\bibitem{selvaraju2017grad}
R.~R. Selvaraju, M.~Cogswell, A.~Das, R.~Vedantam, D.~Parikh, and D.~Batra,
  ``Grad-cam: Visual explanations from deep networks via gradient-based
  localization,'' in \emph{Proceedings of the IEEE international conference on
  computer vision}, 2017, pp. 618--626.

\bibitem{shi2020large}
F.~Shi, L.~Xia, F.~Shan, D.~Wu, Y.~Wei, H.~Yuan, H.~Jiang, Y.~Gao, H.~Sui, and
  D.~Shen, ``Large-scale screening of covid-19 from community acquired
  pneumonia using infection size-aware classification,'' \emph{arXiv preprint
  arXiv:2003.09860}, 2020.

\bibitem{franquet2018imaging}
T.~Franquet, ``Imaging of community-acquired pneumonia,'' \emph{Journal of
  thoracic imaging}, vol.~33, no.~5, pp. 282--294, 2018.

\bibitem{rajpurkar2017chexnet}
P.~Rajpurkar, J.~Irvin, K.~Zhu, B.~Yang, H.~Mehta, T.~Duan, D.~Ding, A.~Bagul,
  C.~Langlotz, K.~Shpanskaya \emph{et~al.}, ``Chexnet: Radiologist-level
  pneumonia detection on chest x-rays with deep learning,'' \emph{arXiv
  preprint arXiv:1711.05225}, 2017.

\bibitem{challenge2018radiological}
R.~P.~D. Challenge, ``Radiological society of north america,'' 2018.

\bibitem{lin2017focal}
T.-Y. Lin, P.~Goyal, R.~Girshick, K.~He, and P.~Doll{\'a}r, ``Focal loss for
  dense object detection,'' in \emph{Proceedings of the IEEE international
  conference on computer vision}, 2017, pp. 2980--2988.

\bibitem{he2017mask}
K.~He, G.~Gkioxari, P.~Doll{\'a}r, and R.~Girshick, ``Mask r-cnn,'' in
  \emph{Proceedings of the IEEE international conference on computer vision},
  2017, pp. 2961--2969.

\bibitem{wielputz2014radiological}
M.~O. Wielp{\"u}tz, C.~P. Heu{\ss}el, F.~J. Herth, and H.-U. Kauczor,
  ``Radiological diagnosis in lung disease: factoring treatment options into
  the choice of diagnostic modality,'' \emph{Deutsches {\"A}rzteblatt
  International}, vol. 111, no.~11, p. 181, 2014.

\bibitem{depeursinge2015automated}
A.~Depeursinge, A.~S. Chin, A.~N. Leung, D.~Terrone, M.~Bristow, G.~Rosen, and
  D.~L. Rubin, ``Automated classification of usual interstitial pneumonia using
  regional volumetric texture analysis in high-resolution ct,''
  \emph{Investigative radiology}, vol.~50, no.~4, p. 261, 2015.

\bibitem{macmahon2017guidelines}
H.~MacMahon, D.~P. Naidich, J.~M. Goo, K.~S. Lee, A.~N. Leung, J.~R. Mayo,
  A.~C. Mehta, Y.~Ohno, C.~A. Powell, M.~Prokop \emph{et~al.}, ``Guidelines for
  management of incidental pulmonary nodules detected on ct images: from the
  fleischner society 2017,'' \emph{Radiology}, vol. 284, no.~1, pp. 228--243,
  2017.

\bibitem{wang2020deep}
S.~Wang, B.~Kang, J.~Ma, X.~Zeng, M.~Xiao, J.~Guo, M.~Cai, J.~Yang, Y.~Li,
  X.~Meng \emph{et~al.}, ``A deep learning algorithm using ct images to screen
  for corona virus disease (covid-19),'' \emph{medRxiv}, 2020.

\bibitem{xu2020deep}
X.~Xu, X.~Jiang, C.~Ma, P.~Du, X.~Li, S.~Lv, L.~Yu, Y.~Chen, J.~Su, G.~Lang
  \emph{et~al.}, ``Deep learning system to screen coronavirus disease 2019
  pneumonia,'' \emph{arXiv preprint arXiv:2002.09334}, 2020.

\bibitem{song2020deep}
Y.~Song, S.~Zheng, L.~Li, X.~Zhang, X.~Zhang, Z.~Huang, J.~Chen, H.~Zhao,
  Y.~Jie, R.~Wang \emph{et~al.}, ``Deep learning enables accurate diagnosis of
  novel coronavirus (covid-19) with ct images,'' \emph{medRxiv}, 2020.

\bibitem{alom2018recurrent}
M.~Z. Alom, M.~Hasan, C.~Yakopcic, T.~M. Taha, and V.~K. Asari, ``Recurrent
  residual convolutional neural network based on u-net (r2u-net) for medical
  image segmentation,'' \emph{arXiv preprint arXiv:1802.06955}, 2018.

\bibitem{9069255}
F.~{Shi}, J.~{Wang}, J.~{Shi}, Z.~{Wu}, Q.~{Wang}, Z.~{Tang}, K.~{He},
  Y.~{Shi}, and D.~{Shen}, ``Review of artificial intelligence techniques in
  imaging data acquisition, segmentation and diagnosis for covid-19,''
  \emph{IEEE Reviews in Biomedical Engineering}, 2020.

\bibitem{hara2018can}
K.~Hara, H.~Kataoka, and Y.~Satoh, ``Can spatiotemporal 3d cnns retrace the
  history of 2d cnns and imagenet?'' in \emph{Proceedings of the IEEE
  conference on Computer Vision and Pattern Recognition}, 2018, pp. 6546--6555.

\bibitem{van2017devil}
G.~Van~Horn and P.~Perona, ``The devil is in the tails: Fine-grained
  classification in the wild,'' \emph{arXiv preprint arXiv:1709.01450}, 2017.

\bibitem{zhou2019bbn}
B.~Zhou, Q.~Cui, X.-S. Wei, and Z.-M. Chen, ``Bbn: Bilateral-branch network
  with cumulative learning for long-tailed visual recognition,'' \emph{arXiv
  preprint arXiv:1912.02413}, 2019.

\bibitem{buda2018systematic}
M.~Buda, A.~Maki, and M.~A. Mazurowski, ``A systematic study of the class
  imbalance problem in convolutional neural networks,'' \emph{Neural Networks},
  vol. 106, pp. 249--259, 2018.

\bibitem{shen2016relay}
L.~Shen, Z.~Lin, and Q.~Huang, ``Relay backpropagation for effective learning
  of deep convolutional neural networks,'' in \emph{European conference on
  computer vision}.\hskip 1em plus 0.5em minus 0.4em\relax Springer, 2016, pp.
  467--482.

\bibitem{he2009learning}
H.~He and E.~A. Garcia, ``Learning from imbalanced data,'' \emph{IEEE
  Transactions on knowledge and data engineering}, vol.~21, no.~9, pp.
  1263--1284, 2009.

\bibitem{japkowicz2002class}
N.~Japkowicz and S.~Stephen, ``The class imbalance problem: A systematic
  study,'' \emph{Intelligent data analysis}, vol.~6, no.~5, pp. 429--449, 2002.

\bibitem{cui2019class}
Y.~Cui, M.~Jia, T.-Y. Lin, Y.~Song, and S.~Belongie, ``Class-balanced loss
  based on effective number of samples,'' in \emph{Proceedings of the IEEE
  Conference on Computer Vision and Pattern Recognition}, 2019, pp. 9268--9277.

\bibitem{chawla2002smote}
N.~V. Chawla, K.~W. Bowyer, L.~O. Hall, and W.~P. Kegelmeyer, ``Smote:
  synthetic minority over-sampling technique,'' \emph{Journal of artificial
  intelligence research}, vol.~16, pp. 321--357, 2002.

\bibitem{wang2018non}
X.~Wang, R.~Girshick, A.~Gupta, and K.~He, ``Non-local neural networks,'' in
  \emph{Proceedings of the IEEE conference on computer vision and pattern
  recognition}, 2018, pp. 7794--7803.

\bibitem{fu2019dual}
J.~Fu, J.~Liu, H.~Tian, Y.~Li, Y.~Bao, Z.~Fang, and H.~Lu, ``Dual attention
  network for scene segmentation,'' in \emph{Proceedings of the IEEE Conference
  on Computer Vision and Pattern Recognition}, 2019, pp. 3146--3154.

\bibitem{hu2018squeeze}
J.~Hu, L.~Shen, and G.~Sun, ``Squeeze-and-excitation networks,'' in
  \emph{Proceedings of the IEEE conference on computer vision and pattern
  recognition}, 2018, pp. 7132--7141.

\bibitem{fukui2019attention}
H.~Fukui, T.~Hirakawa, T.~Yamashita, and H.~Fujiyoshi, ``Attention branch
  network: Learning of attention mechanism for visual explanation,'' in
  \emph{Proceedings of the IEEE Conference on Computer Vision and Pattern
  Recognition}, 2019, pp. 10\,705--10\,714.

\bibitem{li2018tell}
K.~Li, Z.~Wu, K.-C. Peng, J.~Ernst, and Y.~Fu, ``Tell me where to look: Guided
  attention inference network,'' in \emph{Proceedings of the IEEE Conference on
  Computer Vision and Pattern Recognition}, 2018, pp. 9215--9223.

\bibitem{milletari2016v}
F.~Milletari, N.~Navab, and S.-A. Ahmadi, ``V-net: Fully convolutional neural
  networks for volumetric medical image segmentation,'' in \emph{2016 Fourth
  International Conference on 3D Vision (3DV)}.\hskip 1em plus 0.5em minus
  0.4em\relax IEEE, 2016, pp. 565--571.

\bibitem{lin2013network}
M.~Lin, Q.~Chen, and S.~Yan, ``Network in network,'' \emph{arXiv preprint
  arXiv:1312.4400}, 2013.

\bibitem{nair2010rectified}
V.~Nair and G.~E. Hinton, ``Rectified linear units improve restricted boltzmann
  machines,'' in \emph{Proceedings of the 27th international conference on
  machine learning (ICML-10)}, 2010, pp. 807--814.

\bibitem{paszke2019pytorch}
A.~Paszke, S.~Gross, F.~Massa, A.~Lerer, J.~Bradbury, G.~Chanan, T.~Killeen,
  Z.~Lin, N.~Gimelshein, L.~Antiga \emph{et~al.}, ``Pytorch: An imperative
  style, high-performance deep learning library,'' in \emph{Advances in Neural
  Information Processing Systems}, 2019, pp. 8024--8035.

\bibitem{kingma2014adam}
D.~P. Kingma and J.~Ba, ``Adam: A method for stochastic optimization,''
  \emph{arXiv preprint arXiv:1412.6980}, 2014.

\bibitem{xue2006classic}
Z.~Xue, D.~Shen, and C.~Davatzikos, ``Classic: consistent longitudinal
  alignment and segmentation for serial image computing,'' \emph{NeuroImage},
  vol.~30, no.~2, pp. 388--399, 2006.

\end{thebibliography}
\bibliographystyle{IEEEtran}

%








\end{document}